%% file: egpaper_for_review.tex
\documentclass[10pt,twocolumn,letterpaper]{article}

\usepackage{iccv}
\usepackage{times}
\usepackage{epsfig}
\usepackage{graphicx}
\usepackage{amsmath}
\usepackage{amssymb}
\usepackage{bbm}
\usepackage{booktabs}
\usepackage{bm}
\usepackage{multirow}
\usepackage{microtype}
\usepackage[T1]{fontenc}

\usepackage{xcolor}
\usepackage{listings}
\usepackage{paralist}

\newcommand{\NAME}{\textsc{PromptCap}\xspace}

\usepackage[pagebackref=true,breaklinks=true,letterpaper=true,colorlinks,bookmarks=false]{hyperref}

\iccvfinalcopy 

\def\iccvPaperID{8675} 

\ificcvfinal\fi

\begin{document}

\title{\NAME: Prompt-Guided Task-Aware Image Captioning}

\author{
     \textbf{Yushi Hu}$^{1}\thanks{Equal contribution. Correspondance to <Yushi Hu: yushihu@uw.edu>, <Hang Hua: hhua2@cs.rochester.edu>}$ \quad
     \textbf{Hang Hua}$^{2*}$\quad
     \textbf{Zhengyuan Yang}$^{3}$\quad \\
     \textbf{Weijia Shi}$^{1}$\quad
     \textbf{Noah A. Smith}$^{1,4}$\quad
     \textbf{Jiebo Luo}$^{2}$\quad\\
     $^1$University of Washington\quad
     $^2$University of Rochester\\
     $^3$Microsoft\quad
     $^4$Allen Institute for AI\\
     {\tt \textcolor{pink}{\url{https://yushi-hu.github.io/promptcap_demo/}}}
}

\maketitle
\ificcvfinal\thispagestyle{empty}\fi

\begin{abstract}
\input{text/abstract}
\end{abstract}

\input{text/intro}
\input{text/realtedwork}

\input{text/approach}

\input{text/approach2.tex}
\input{text/experiments.tex}

\input{text/discussion.tex}
\input{text/conclusion.tex}

{\small
\bibliographystyle{ieee_fullname}
\bibliography{references}
}

\clearpage
\appendix
\input{text/appendix}

\end{document}

%% file: text/abstract.tex
Knowledge-based visual question answering (VQA) involves questions that require world knowledge beyond the image to yield the correct answer. Large language models (LMs) like GPT-3 are particularly helpful for this task because of their strong knowledge retrieval and reasoning capabilities. 
To enable LM to understand images, prior work uses a captioning model to convert images into text.
However, when summarizing an image in a single caption sentence, which visual entities to describe are often underspecified. 
Generic image captions often miss visual details essential for the LM to answer visual questions correctly.
To address this challenge, we propose \NAME (\textbf{Prompt}-guided image \textbf{Cap}tioning), a captioning model designed to serve as a better connector between images and black-box LMs.
Different from generic captions, \NAME takes a natural-language prompt to control the visual entities to describe in the generated caption. The prompt contains a question that the caption should aid in answering.
To avoid extra annotation, \NAME is trained by examples synthesized with GPT-3 and existing datasets.
We demonstrate \NAME's effectiveness on an existing pipeline in which GPT-3 is prompted with image captions to carry out VQA.
\NAME outperforms generic captions by a large margin and achieves state-of-the-art accuracy on knowledge-based VQA tasks (60.4\% on OK-VQA and 59.6\% on A-OKVQA).
Zero-shot results on WebQA show that \NAME generalizes well to unseen domains.\footnote{All codes, data, and demos are available on the project page. HF checkpoint: \url{https://huggingface.co/tifa-benchmark/promptcap-coco-vqa}}


%% file: text/intro.tex
\section{Introduction}
\label{sec:intro}

\begin{figure}[ht]
\centering
  \includegraphics[width=0.45\textwidth]{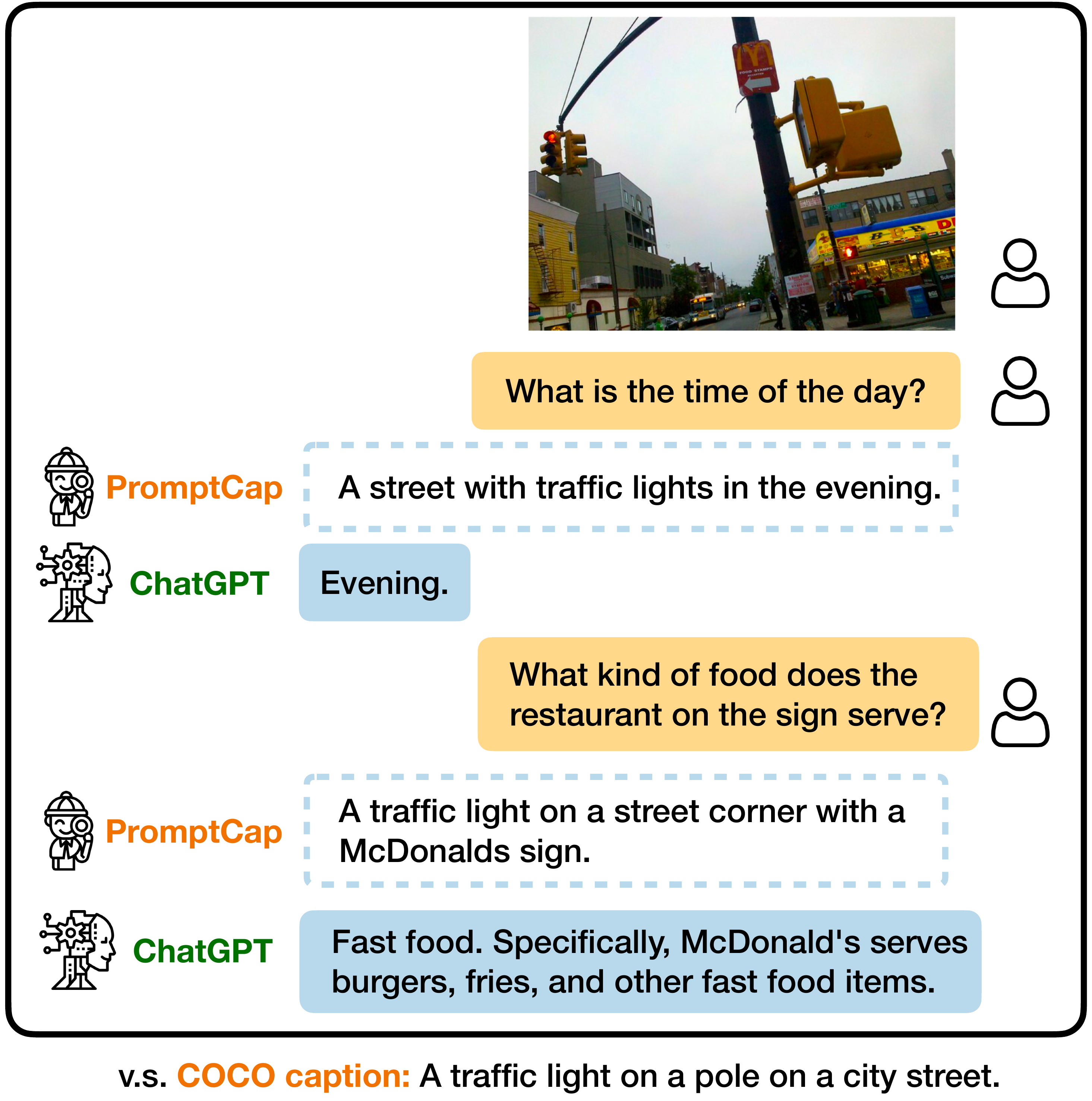}
  \caption{
  Illustration of VQA with \NAME and ChatGPT.
  \NAME is designed to work with black-box language models (\eg, GPT-3, ChatGPT) by describing question-related visual information in the text. Different from generic captions, \NAME customizes the caption according to the input question prompt, which helps ChatGPT understand the image and give correct answers to the user. In contrast, ChatGPT cannot infer the answers from the vanilla human-written caption from MSCOCO. 
}
  \vspace{-5mm}
  \label{fig:teaser}
\end{figure}

Knowledge-based visual question answering (VQA) \cite{marino2019ok} extends traditional VQA tasks~\cite{antol2015vqa} with questions that require broad knowledge and commonsense reasoning to yield the correct answer. 
Existing systems on knowledge-based VQA retrieve external knowledge from various sources, including knowledge graphs~\cite{garderes2020conceptbert, marino2021krisp, wu2022multi}, Wikipedia~\cite{marino2021krisp, wu2022multi, Gao_2022_CVPR, gui2022kat, Lin2022REVIVERV}, and web search~\cite{luo2021weakly, wu2022multi}. 
Recent work~\cite{yang2022empirical} finds that modern language models (LMs) like GPT-3~\cite{brown2020language} are particularly useful for this task because of their striking knowledge retrieval and reasoning abilities. The current state-of-the-art methods~\cite{yang2022empirical, gui2022kat, Lin2022REVIVERV, alayrac2022flamingo} all make use of recent large language models (GPT-3 or Chinchilla).

One key challenge is to allow LMs to understand images. 
Many top-performing LMs (\eg, GPT-3, ChatGPT) are only accessible via APIs, making it impossible to access their internal representations or conduct fine-tuning~\cite{Shi2023REPLUGRB}. A popular solution is to project images into texts that black-box LMs can process, via a generic image captioning model~\cite{chen2015microsoft} or an image tagger~\cite{yang2022empirical}. This framework has been successful on multiple tasks, including VQA~\cite{yang2022empirical, gui2022kat, Lin2022REVIVERV}, image paragraph captioning~\cite{xie2022visual}, and video-language tasks~\cite{zeng2022socratic,wang2022language}. 
Despite promising results, converting visual inputs into a generic, finite text description risks excluding information necessary for the task. As discussed in PICa~\cite{yang2022empirical}, when used for VQA tasks, the generic caption might miss the detailed visual information needed to answer the question, such as missing the ``McDonald's" in Figure~\ref{fig:teaser}.

\begin{figure}[t]
\centering
  \includegraphics[width=0.48\textwidth]{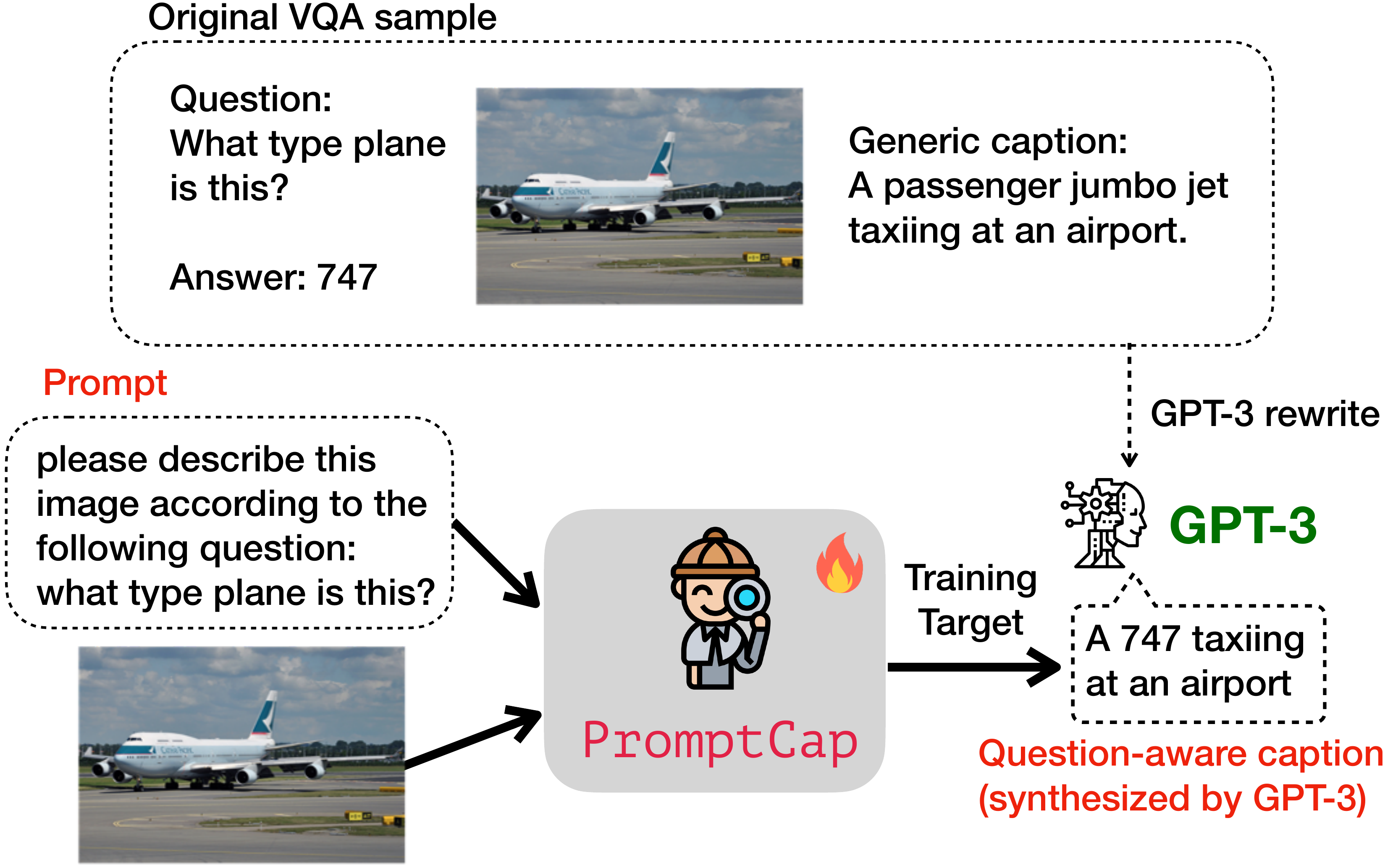}
  \caption{
Overview of \NAME training. \NAME takes two inputs, including an image and a natural language prompt. The model is trained to generate a caption that helps downstream LMs to answer the question. During training, we use GPT-3 to synthesize VQA samples into captioning examples. The original caption is rewritten into a caption that helps answer the question. \NAME is trained to generate this synthesized caption given the image and the prompt.
}

  \label{fig:overview}
\end{figure}

To address the above challenges, we introduce \NAME, a question-aware captioning model designed to serve as a better connector between images and a black-box LM. \NAME is illustrated in Figure~\ref{fig:overview}. \NAME takes an extra natural language prompt as input to control the visual content to describe. The prompt contains the question that the generated caption should help to answer. 
LMs can better answer visual questions by using \NAME as their ``visual front-end".
For example, in Figure~\ref{fig:teaser}, when asked ``what is the time of the day?", \NAME includes ``in the evening" in its image description; when asked ``what kind of food does the restaurant on the sign serve?", \NAME includes ``McDonald's” in its description. Such visual information is critical for ChatGPT to reply to the user with the correct answers. In contrast, the generic COCO~\cite{lin2014microsoft} caption often contains no information about the time or the sign, making ChatGPT unable to answer the questions.

One major technical challenge is \NAME training. The pipeline of ``\NAME + black-box LM" cannot be end-to-end fine-tuned on VQA tasks because the LM parameters are not exposed through the API. Also, there are no training data for question-aware captions. 
To avoid extra annotation, we propose a pipeline to synthesize and filter training samples with GPT-3.
Specifically, we view existing VQA datasets as pairs of question and question-related visual details.
Given a question-answer pair, we rewrite the corresponding image's generic caption into a customized caption that helps answer the question. Following 20 human-annotated examples, GPT-3 synthesizes a large number of question-aware captions via few-shot in-context learning~\cite{brown2020language}. To ensure the sample quality, we filter the generated captions by performing QA with GPT-3, checking if the answer can be inferred given the question and the synthesized caption.
Notice that GPT-3 is frozen in the whole pipeline. Its strong few-shot learning ability makes this pipeline possible.

We demonstrate the effectiveness of \NAME on knowledge-based VQA tasks with the pipeline in PICa~\cite{yang2022empirical}.
Details of the pipeline are illustrated in \S\ref{sec:icl}. The images are converted into texts via \NAME, allowing GPT-3 to perform VQA via in-context learning. This pipeline, despite its simplicity, achieves state-of-the-art results on knowledge-based VQA tasks (\textbf{60.4\%} on OK-VQA~\cite{marino2019okvqa} and \textbf{59.6\%} on A-OKVQA~\cite{schwenk2022aokvqa}). We also conduct extensive ablation studies on the contribution of each component, showing that \NAME gives a consistent performance gain (\textbf{3.8\%} on OK-VQA, \textbf{5.3\%} on A-OKVQA, and \textbf{9.2\%} on VQAv2) over a generic captioning model that shares the same architecture and training data. Finally, we investigate \NAME's generalization ability on WebQA~\cite{chang2022webqa}, showing that \NAME, without any training on the compositional questions in WebQA, outperforms the generic caption approach and all supervised baselines.

In summary, our contributions are as follows:

\begin{itemize}
    \item We propose \NAME, a novel question-aware captioning model that uses natural language prompt to control the visual content to be described. (\S\ref{sec:promptcap})
    \item To the best of our knowledge, we are the first to propose a pipeline to synthesize and filter training samples for vision-language tasks via GPT-3 (\S\ref{sec:approach:example_gen}).
    \item \NAME helps GPT-3 in-context learning (\S\ref{sec:icl}) achieve state-of-the-art results on OK-VQA and A-OKVQA, substantially outperforming generic captions on various VQA tasks. (\S\ref{sec:experiment}).

\end{itemize}


%% file: text/realtedwork.tex
\section{Related Work}
\paragraph{Knowledge-Based VQA}
Knowledge-based VQA \cite{marino2019okvqa,schwenk2022aokvqa} requires systems to leverage external knowledge beyond image content to answer the question. 
Prior works  \cite{garderes2020conceptbert, marino2021krisp, wu2022multi, zhu2020mucko,narasimhan2018out,narasimhan2018straight, izacard2020distilling, izacard2020leveraging, Gao_2022_CVPR} investigate leveraging knowledge from various external knowledge resources, e.g., Wikipedia \cite{vrandevcic2014wikidata}, ConceptNet \cite{speer2017conceptnet}, and ASER \cite{zhang2020aser}, to improve the performance of the VQA models. 
Inspired by PICa~\cite{yang2022empirical}, recent works~\cite{gui2022kat, Lin2022REVIVERV} use GPT-3 as an implicit knowledge base and achieve state-of-the-art results. We identify the critical problem:  generic captions used to prompt GPT-3 often miss critical visual details for VQA. We address this challenge with \NAME.

\vspace{-0.1in}
\paragraph{Vision-Language Models}
Vision-language models have recently shown striking success on various multimodal tasks~\cite{su2019vl,lu2019vilbert,chen2019uniter,li2020oscar,wang2021simvlm,kim2021vilt, radford2021learning,yang2022unitab,wang2022git,wang2022ofa, Lu2022UnifiedIOAU,yuan2021florence,chen2022pali, Li2023BLIP2BL}. These works first pretrain multimodal models on large-scale image-text datasets and then finetune the models for particular tasks. The works most related to ours are Frozen~\cite{tsimpoukelli2021multimodal}, Flamingo~\cite{alayrac2022flamingo}, and BLIP-2~\cite{Li2023BLIP2BL}, which keeps the LMs frozen and tune a visual encoder for the LM. However, such techniques require access to internal LM parameters and are thus difficult to be applied to black-box LMs like GPT-3. 

\vspace{-0.1in}
\paragraph{Prompting for Language Models}
Prompting allows a pre-trained model to adapt to different tasks via different prompts without modifying any parameters. LLMs like GPT-3~\cite{brown2020language} have shown strong zero-shot and few-shot ability via prompting.
Prompting has been successful for a variety of natural language tasks~\cite{liu2021pre}, including but not limited to  classification tasks~\cite{min2022metaicl, shi2022nearest}, semantic parsing~\cite{xie2022unifiedskg}, knowledge generation~\cite{Shi2023REPLUGRB, liu2022generated}, and dialogue systems~\cite{lee2021dialogue, hu-etal-2022-context}.
The most closely-related works to ours are the instruction-finetuned language models~\cite{t0,wei2021finetuned, Wang2022SuperNaturalInstructionsGV}.


%% file: text/approach.tex
\section{\NAME}
\label{sec:promptcap}

We introduce \NAME, an image captioning model that utilizes a natural language prompt as an input condition.
The overview of \NAME training is in Figure~\ref{fig:overview}.
Given an image $I$, and a natural language prompt $P$, \NAME generates a prompt-guided caption $C$.
$P$ contains instructions about the image contents of interest to the user. 
For VQA, an example prompt could be \textit{``Please describe this image according to the following question: what type plane is this?}.
The prompt-guided caption $C$ should (1) cover the visual details required by the instruction in the prompt, (2) describe the main objects as general captions do, and (3) use auxiliary information in the prompt if necessary.
For instance, assuming the prompt contains a VQA question, $C$ may directly describe the asked visual contents (\eg, for questions about visual details), or provide information that helps downstream models to infer the answer (\eg, for questions that need external knowledge to solve). 

Given the above design, the major technical challenge is \NAME training. \NAME is designed to work with black-box LMs, which cannot be end-to-end fine-tuned on VQA tasks because the LM parameters are not accessible. Besides, there are no training data for question-aware captions. To address these challenges, we propose training \NAME with data synthesized with GPT-3.

\subsection{Training Data Synthesis}
\label{sec:approach:example_gen}

To avoid annotating question-aware caption examples, we use GPT-3 to generate training examples for \NAME via in-context learning~\cite{brown2020language, rubin2022learning, hu-etal-2022-context, cheng2022binding}.

\vspace{-0.1in}
\subsubsection{Training Example Generation with GPT-3}

\begin{figure}[t]
\centering
  \includegraphics[width=0.45\textwidth]{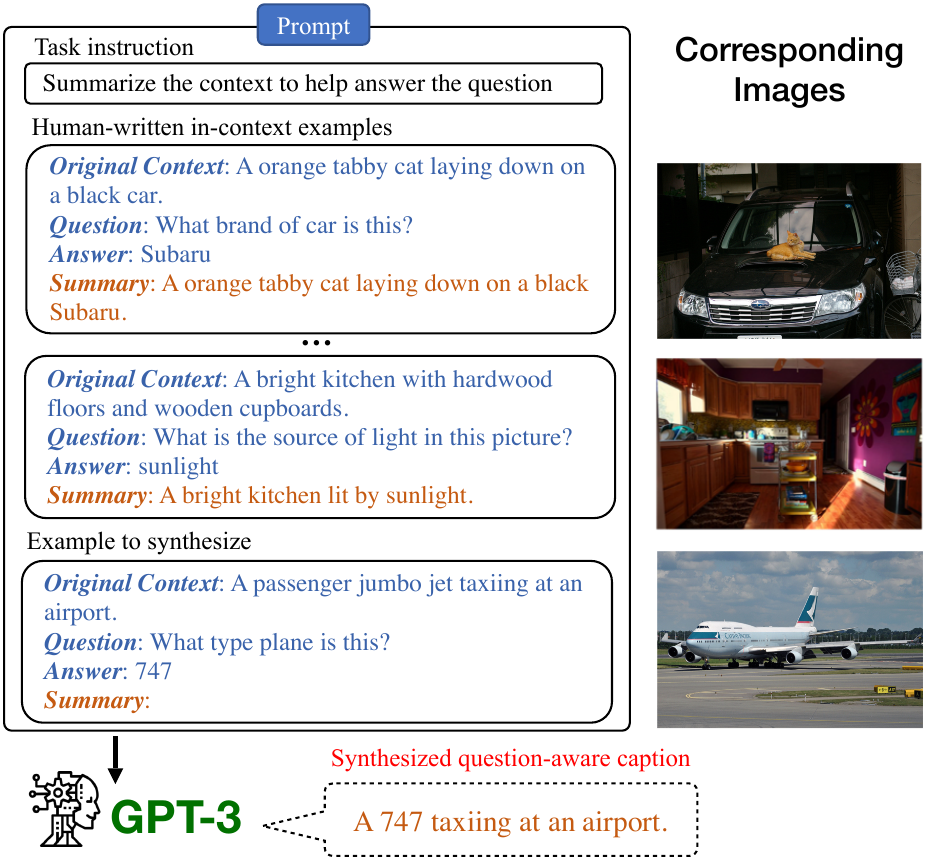}
  \caption{
Training example synthesis with GPT-3 in-context learning.
The ``Original Contexts" are ground-truth image captions.
The question-answer pairs come from existing VQA datasets.
GPT-3 generalizes (without parameter updates) from the human-written examples to produce the question-aware caption given the caption, question, and answer.  The images are shown for clarity but are {\it not} used in our data synthesis procedure. 
}
  \vspace{-4mm}
  \label{fig:example_gen}
\end{figure}

For \NAME training, we view existing VQA datasets as natural sources of pairs of task and task-related visual details.
We synthesize question-aware captions by combining the general image captions and the question-answering pairs using GPT-3 in-context learning.
Figure~\ref{fig:example_gen} illustrates the GPT-3 prompt we use for training example generation.
The prompt contains the task instruction, 20 human-written examples, and the VQA question-image pair that we synthesize the task-aware caption from.
Since GPT-3 only accepts text inputs, we represent each image by concatenating the 5 human-written COCO captions~\cite{chen2015microsoft}, as shown in the ``Original Context". 
The human-written examples follow the three principles of prompt-guided captions described in Section~\ref{sec:promptcap}.
The commonsense reasoning ability of GPT-3 allows the model to understand the image to some extent via the COCO captions and synthesize new examples by following the human-written examples.

\vspace{-0.1in}
\subsubsection{Training Example Filtering}
To ensure the quality of the generated captions, we sample 5 candidate captions from GPT-3 for each question-answer pair.
We devise a pipeline to filter out the best candidate caption as the training example for \NAME.
The idea is that a text-only 
QA system should correctly answer the question given a high-quality prompt-guided caption as the context.
For each candidate caption, we use the GPT-3 in-context learning VQA system in \S\ref{sec:icl} to predict an answer, and score the candidate captions by comparing this answer with the ground-truth answers.

\vspace{-0.1in}
\paragraph{Soft VQA Accuracy} We find that in the open-ended generation setting, the VQA accuracy~\cite{goyal2017making} incorrectly punishes answers with a slight difference in surface form. 
For example, the answer ``coins" gets $0$ when the ground truth is ``coin".
To address this problem, we devise a new soft VQA accuracy for example filtering.
Suppose the predicted answer is $a$ and the human-written ground truth answers are $[g_1, g_2, ..., g_n]$. The soft accuracy is given by 
the three lowest-CER ground truth answers:
\begin{equation*}
    \mathit{Acc}_{\mathit{soft}}(a) = \underset{x,y,z \in [n]}{\max}{\sum_{i \in \{x,y,z\}}\frac{\max[0, 1-\mathit{CER}(a, g_i)]}{3}},
\end{equation*}
where CER is the character error rate, calculated by the character edit distance over the total number of characters of the ground truth. 
In contrast, the traditional VQA accuracy~\cite{goyal2017making} uses exact match.
We sort the candidate captions based on this soft score.

\vspace{-0.1in}
\paragraph{Comparing with COCO ground-truth}
Multiple candidates may answer the question correctly and get the same soft score.
To break ties, we also compute the CIDEr score~\cite{vedantam2015cider} between the candidate captions and the COCO ground-truth captions.
Among the candidates with the highest soft VQA accuracy, the one with the highest CIDEr score is selected as the training example for \NAME.

\subsection{\NAME Training}
For \NAME training, we start with the state-of-the-art pre-trained vision-language model OFA \cite{wang2022ofa} and make some modifications to the OFA captioning model. 
OFA has an encoder-decoder structure.
As discussed earlier, our training data are synthesized with VQA data in the form of question-caption pairs.
Given a question-caption pair,
we first rewrite the question into an instruction prompt via a template.
For example, the instruction prompt might be
\textit{``describe to answer: What is the clock saying the time is?"}.
We apply byte-pair encoding (BPE) \cite{sennrich2015neural} to the given text sequence, encoding it as subwords.
Images are transformed into image patches that share the same subword token set.
Let the training samples be $\mathcal{D} = \{P_i, I_i, C_i\}_{i=1}^{|\mathcal{D}|}$, in which $P_i$ is the text prompt, $I_i$ is the image patch, and $C_i$ is the synthesized task-aware caption.
The captioning model takes $[P_i:I_i]$ as input and is trained to generate $C_i=[c_1, c_2, ... , c_{|C_i|}]$. Here $[:]$ is concatenation. We use negative log-likelihood loss and train the model in an end-to-end manner. The training loss is :
\begin{equation*}
\mathcal{L} =- \sum_{\mathcal{D}} \sum_{t=1}^{|C_i|} \log p (c_t\mid [P_i:I_i], c_{\leq t-1}).
\end{equation*}


%% file: text/approach2.tex
\begin{figure*}[h]
\centering
  \includegraphics[width=0.9\textwidth]{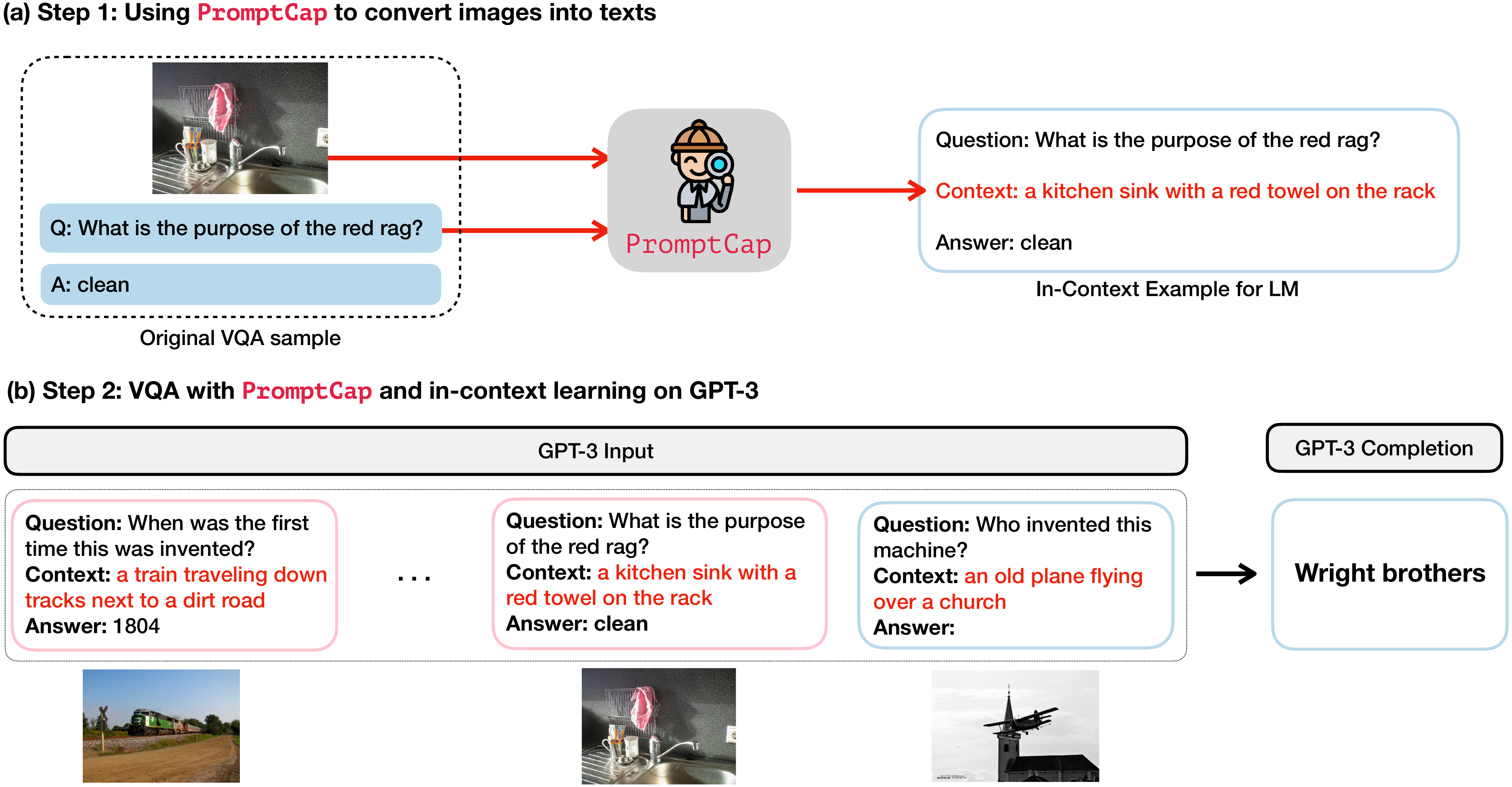}
  \caption{
Our inference pipeline for VQA. \textbf{(a) Illustration of how we convert a VQA sample into pure text.} Given the image and the question, \NAME describes the question-related visual information in natural language. The VQA sample is turned into a QA sample that GPT-3 can understand. 
\textbf{(b) GPT-3 in-context learning for VQA.} After converting the VQA examples into text with \NAME, we carry out VQA by in-context learning on GPT-3. The input consists of the task instruction (not shown in the figure), the in-context examples, and the test instance. GPT-3 takes the input and generates the answer. Notice that the GPT-3 is treated as a black box and is only used for inference. The question-aware captions \NAME generated are marked \textcolor{red}{red}.
}
  \vspace{-4mm}
  \label{fig:end_task}
\end{figure*}

\section{VQA with \NAME and GPT-3}
\label{sec:icl}

Our VQA pipeline is illustrated in Figure~\ref{fig:end_task}, which is adopted from PICa~\cite{yang2022empirical}.
The pipeline consists of two components, \NAME and GPT-3. 

\vspace{-0.1in}
\paragraph{Step 1: Converting images into texts via \NAME}
GPT-3 can perform a new task by simply conditioning on several task training examples as demonstrations. As we have discussed, the major challenge is that GPT-3 does not understand images. To bridge this modality gap, we convert the images in VQA samples to texts using \NAME (Figure~\ref{fig:end_task}a). Notice that different from generic captioning models, \NAME customizes the image caption according to the question, which enables LMs to understand question-related visual information in the image.
As such, we are able to convert VQA samples into question-answering examples that GPT-3 can understand.

\vspace{-0.1in}
\paragraph{Step 2: GPT-3 in-context learning for VQA}
Having used \NAME to convert VQA examples into question-answer examples that GPT-3 can understand (Step 1), 
we use a subset of these examples as the task demonstration for GPT-3. We concatenate the in-context learning examples to form a prompt, as shown in Figure~\ref{fig:end_task}b. Each in-context learning example consists of a question (\texttt{Question: When was the first time this was invented?}), a context generated by \NAME (\texttt{Context: a train traveling down tracks next to a dirt road}), and an answer (\texttt{Answer: 1804}). Then we append the test example to the in-context learning examples, and provide them as inputs to GPT-3. GPT-3 generates predictions based on an open-ended text generation approach, taking into account the information provided in the in-context learning examples and the test example.

\vspace{-0.1in}
\paragraph{Example retrieval}
\label{sec:icl:retrieval}
Previous research has shown that the effectiveness of in-context learning examples chosen for GPT-3 can significantly impact its performance \cite{liu-etal-2022-makes}. In the few-shot setting where only a few training examples are available, we simply use these examples as in-context learning examples (referred to as ``Random'' in later sections because they are selected at random from our collection). 
However, in practice, we often have access to more than small-$n$ examples (i.e., full training data setting). To improve the selection of in-context learning examples, we follow the approach proposed by \cite{yang2022empirical}:  we compute the similarity between examples using CLIP \cite{radford2021learning} by summing up the cosine similarities of the question and image embeddings. The $n$ most similar examples in the training set are then selected as in-context examples (referred to as ``CLIP'' in this paper). By using the most similar in-context examples to the test instance, our approach can improve the quality of the learned representations and boost the performance of GPT-3 on VQA tasks.

%% file: text/experiments.tex
\input{tables/okvqa}
\input{tables/aokvqa}

\section{Experiments}
\label{sec:experiment}

In this section, we demonstrate \NAME's effectiveness on knowledge-based VQA tasks.
First, we show that \NAME captions enable GPT-3 to achieve state-of-the-art performance on OK-VQA~\cite{marino2019okvqa} and A-OKVQA~\cite{schwenk2022aokvqa} with in-context learning.
Then we conduct ablation experiments on the contribution of each component, showing that \NAME is giving consistent gains over generic captions.
In addition, experiments on WebQA~\cite{chang2022webqa} demonstrate that \NAME generalizes well to unseen domains.

\subsection{Experimental Setup}
\label{sec:results:setup}

\paragraph{Datasets} We use three knowledge-based VQA datasets, namely OK-VQA~\cite{marino2019okvqa}, A-OKVQA~\cite{schwenk2022aokvqa}, and WebQA~\cite{chang2022webqa}.
\textbf{OK-VQA}\cite{marino2019okvqa} is a large knowledge-based VQA dataset that contains 14K image-question pairs. Questions are manually filtered to ensure that outside knowledge is required to answer the questions.
\textbf{A-OKVQA}\cite{schwenk2022aokvqa} is an augmented successor of OK-VQA, containing 25K image-question pairs that require broader commonsense and world knowledge to answer.
For both OK-VQA and A-OKVQA, the direct answers are evaluated by the soft accuracy from VQAv2\cite{goyal2017making}.
Besides direct answer evaluation,
A-OKVQA also provides multiple-choice evaluation, where the model should choose one correct answer among 4 candidates. 
\textbf{WebQA} \cite{chang2022webqa} is a multimodal multi-hop reasoning benchmark that requires the model to combine multiple text and image sources to answer a question.

\vspace{-0.1in}
\paragraph{\NAME implementation details}
We adopt the officially released OFA~\cite{wang2022ofa} captioning checkpoint ``caption-large-best-clean'' (470M) for model initialization and use the GPT-3 synthesized examples in \S\ref{sec:approach:example_gen} to fine-tune the model. The examples are synthesized from VQAv2~\cite{antol2015vqa, goyal2017making}. Notice that this dataset is included in OFA pre-training, so we are not adding additional annotated training data compared with OFA. We use AdamW \cite{Kingma2015AdamAM} optimizer with learning rate $ \{2 \times 10^{-5}, 3 \times 10^{-5}, 5 \times 10^{-5} \} $, batch size $ \{32, 64, 128\}$, and $\beta_1=0.9,\ \beta_2=0.999$ for training.

\vspace{-0.1in}
\paragraph{In-context learning details} We use {\it code-davinci-002} engine (175B) for GPT-3 in all the experiments. Due to the input length limit, we use $n=32$ most similar examples in the prompt for GPT-3. The examples are retrieved by CLIP (VIT-L/14) using the method discussed in \S\ref{sec:icl:retrieval}.

\subsection{Results on OK-VQA and A-OKVQA}

Table 1 compares \NAME + GPT-3 with other methods on the OK-VQA validation set. For each method, we also list the way it represents the images, and the knowledge source used. The table is split into two sections. The upper section lists fully supervised methods. These methods require end-to-end finetuning. The methods in the bottom section are based on in-context learning and no task-specific finetuning is done on the models. 

We can see that all state-of-the-art systems use GPT-3 (or Chinchilla) as part of their systems. These methods obtain significant performance gains compared with previous methods, showing the importance of the LM in knowledge-based VQA tasks. 
PICa is the first system that used GPT-3 as the knowledge source. KAT~\cite{gui2022kat} further improves over PICa by introducing Wikidata~\cite{vrandevcic2014wikidata} as the knowledge source, doing ensemble and end-to-end finetuning on multiple components.
REVIVE~\cite{Lin2022REVIVERV} is the current state of the art on OK-VQA. Compared with KAT, it introduces extra object-centric visual features to the ensemble, which brings additional gains over KAT. However, all of the above methods use generic image captions to prompt knowledge from GPT-3. We identify this as a critical bottleneck in using LMs for VQA tasks. \NAME is designed to address this bottleneck.

\vspace{-0.1in}
\paragraph{Comparison with state of the art} Our proposed \NAME + GPT-3, despite using no additional knowledge source, no ensemble with visual features, and no end-to-end finetuning, achieves \textbf{60.4\%} accuracy and outperforms all existing methods on OK-VQA. Table~\ref{tab:aokvqa} shows similar results on A-OKVQA, in which \NAME + GPT-3 outperforms all prior methods by a large margin on both multiple-choice (\textbf{73.1\%}) and direct-answer (\textbf{59.6\%}) evaluations. These results demonstrate \NAME's effectiveness in connecting LMs with images. Besides, we would like to emphasize that \NAME could replace the captioning module in the systems KAT and REVIVE have proposed, which might further boost the performance. We expect that \NAME will help future systems with complementary advances to achieve even better performance on these tasks.

\subsection{Ablation Study}
\label{sec:exp:ablation}
We conduct extensive ablation studies to quantify the performance benefit of each component in our system,
\ie, the captioning model \NAME, the language model, and the prompting method. We conduct ablation experiments on each component. 

\vspace{-0.1in}
\paragraph{Additional dataset for analysis} Besides knowledge-based VQA tasks, we would also like to investigate the performance gain from \NAME for traditional VQA. 
Thus, we also include VQAv2~\cite{antol2015vqa} in our ablation studies.

\input{tables/ablation}


\vspace{-0.1in}
\subsubsection{Performance Benefit from \NAME}

\paragraph{Baseline generic captioning model} We use the officially released OFA~\cite{wang2022ofa} captioning checkpoint ``caption-large-best-clean'' (470M) as the baseline generic captioning model. We refer to it as ``OFA-Cap''. We choose this model because this is the model initialization we use for \NAME, sharing the same model architecture. Notice that OFA is a large vision-language model pre-trained on 20M image-text pairs and 20 vision-language tasks, including many VQA tasks. We are not using additional annotated data during \NAME finetuning.

\vspace{-0.1in}
\paragraph{\NAME captions give consistent gains over generic captions.}Table~\ref{tab:ablation} measures the performance benefit from \NAME on OK-VQA, A-OKVQA, and VQAv2 validation sets.
Here we focus on the performance gap between using \NAME captions and generic OFA captions.
We can see that \NAME gives consistent improvements over generic captions. 
Specifically, with GPT-3, \NAME improves over OFA-Cap by \textbf{3.8\%}, \textbf{5.3\%}, and \textbf{9.2\%} absolute accuracy on OK-VQA, A-OKVQA, and VQAv2, respectively.

\input{tables/ablation_GPT3}

\vspace{-0.1in}
\subsubsection{Performance Benefit from Language Model}

\paragraph{Baseline language model} To measure the performance gain from GPT-3, we choose Flan-T5-XXL(11B)~\cite{Chung2022ScalingIL} as the baseline language model. FlanT5-XXL is an instruction-finetuned LM that has shown good in-context learning ability. Notice that for Flan-T5-XXL, because of the input length limit, we use $n=16$ in-context examples in the input.

\paragraph{GPT-3 yields huge gains on knowledge-based VQA, but not on VQAv2.} Results in Table~\ref{tab:ablation_gpt3} quantify the benefit of GPT-3 over Flan-T5-XXL. GPT-3 yields great performance gains on knowledge-based VQA tasks, improving over Flan-T5 by \textbf{18.4\%} and \textbf{14.8\%} absolute accuracy on OK-VQA and A-OKVQA, respectively. In comparison, on VQAv2, GPT-3 only gives \textbf{$3.2\%$} accuracy gain, which is much smaller than the gain from \NAME over generic captions. The results indicate that GPT-3's external knowledge is critical for knowledge-based VQA tasks but not for VQAv2. We speculate that this disparity arises from VQAv2's focus on information in the image, without requiring additional knowledge beyond visual information.

\input{tables/ablation_prompting}

\vspace{-0.1in}
\subsubsection{Ablation on GPT-3 Prompting}
As discussed in \S\ref{sec:icl}, two factors affect the in-context learning performance: the number of in-context examples, and the example selection strategy. To measure the effects of these two factors, we conduct an ablation study on GPT-3 prompting for OK-VQA in Table~\ref{tab:prompt_ablation}. We vary the number of examples $n\in \{1,4,16,32\}$ and experiment with random examples and the most similar examples retrieved by CLIP (VIT-L/14)~\cite{radford2021learning}. We can see that for both example selection strategies, the more in-context examples, the better the performance. Also, retrieving most similar examples with CLIP gives substantial performance gain (\textbf{5.2\%} absolute accuracy for \NAME when $n=32$). Both findings agree with the claims in prior work~\cite{brown2020language, liu-etal-2022-makes, yang2022empirical}.

\input{tables/webqa}

\begin{figure*}[h]
\centering
  \includegraphics[width=1\textwidth]{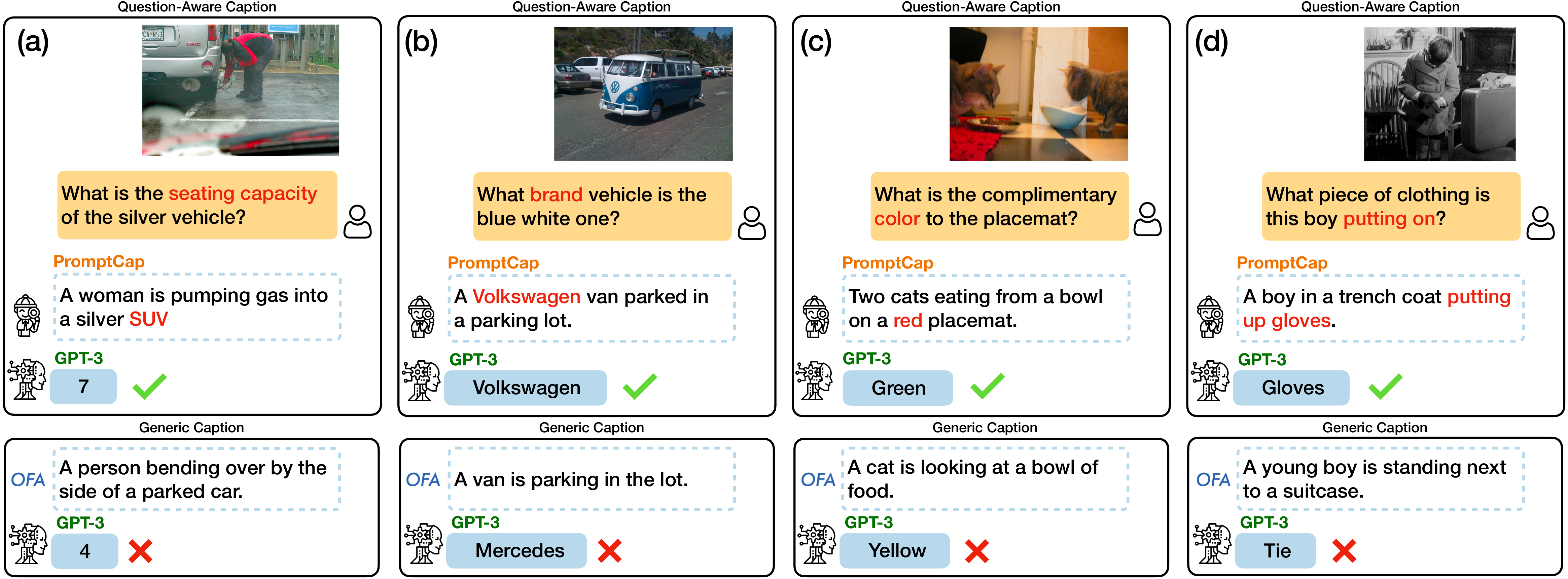}
  
  \caption{Example captions generated by \NAME and OFA-Cap, and the answers GPT-3 generated the captions. For all these questions, GPT-3 yields the correct answer given \NAME captions but fails given the generic caption. Questions are from OK-VQA.
 }

  \label{fig:ok_vqa_demo}
\end{figure*}

\begin{figure*}[h]
\centering
  \includegraphics[width=1\textwidth]{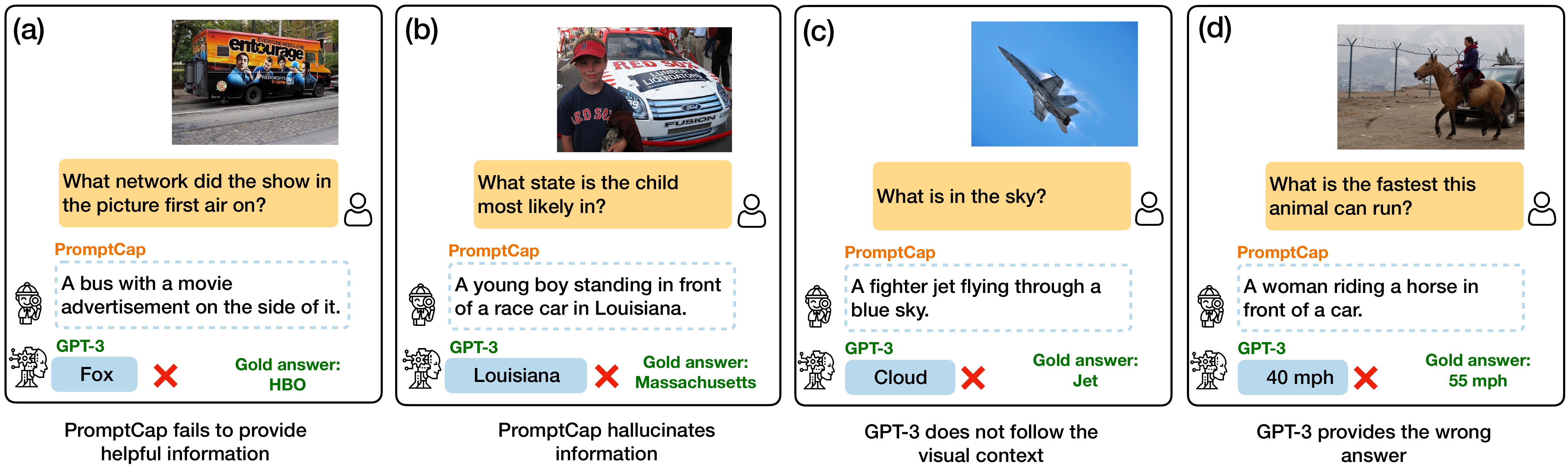}
  
  \caption{Representative failure cases of \NAME and GPT-3 pipeline on OK-VQA.
 }
\vspace{-4mm}
  \label{fig:failures_demo}
\end{figure*}

\subsection{Domain Transfer on WebQA}
We apply \NAME to WebQA~\cite{chang2022webqa} to evaluate \NAME's generalization ability on images and tasks from different domains.
WebQA images are crawled from the web and are from domains different from the COCO~\cite{lin2014microsoft} images used in \NAME's training data synthesized from VQAv2~\cite{goyal2017making}.
Due to the task setting, questions are compositional and much longer than typical VQA questions.
We convert the source images into captions and use GPT-3 in-context learning to carry out the task with only 8 random examples.
The answers are long-form and measured by two scores: the fluency score measured by BARTScore~\cite{yuan2021bartscore} and the accuracy score that measures if human-annotated keywords are included in the answer.
The results in the image-query setting with oracle sources on the validation set\footnote{Setting at \url{https://github.com/WebQnA/WebQA/tree/main/baseline_output_files/Baseline_prediction_files_on_Val}}
are shown on Table~\ref{tab:webqa}.
Our systems outperform all the official baselines.
\NAME outperforms the generic OFA captions, showing that \NAME is generalizable to  a different domain of questions and images.



\subsection{Qualitative Analysis}
Representative captions generated by \NAME and OFA are illustrated in Figure~\ref{fig:ok_vqa_demo}. The task is to answer the questions in OK-VQA. For all these questions, GPT-3 generates the correct answer when taking \NAME's caption as input, but fails when taking the generic caption. \NAME is able to capture visual attributes according to the question, for example, ``brand" in (b) and ``color" in (c). In addition, it can focus on particular objects asked in the question, such as the clothing the boy is ``putting on" in (d). For tasks beyond \NAME's reasoning ability, GPT-3 infers the answer by the visual details \NAME gives. For example, GPT-3 infers ``green" for the ``complimentary color of red" in (c) and ``7" for SUV's seating capacity in (a).
We also show some representative failure cases in Figure~\ref{fig:failures_demo}. The majority of failures are as shown in (a) and (b), in which \NAME fails to provide helpful information, or provides unfactual information. GPT-3 sometimes makes mistakes, as shown in (c) and (d).

\input{tables/captioning}

\subsection{Analysis: How Do Question-Aware Captions Differ from Generic Captions?}
\label{sec:appendix:captioning}

To further analyze the question-aware captions, we compare different inferred/gold captions in Table~\ref{tab:captioning}.
The captions are compared by the automatic evaluations used in MSCOCO~\cite{chen2015microsoft}: BLEU-4 (B)~\cite{papineni2002bleu}, METEOR (M)~\cite{banerjee2005meteor}, CIDEr (C)~\cite{vedantam2015cider}, and SPICE (S)~\cite{anderson2016spice}.
We evaluate the captions on the VQAv2 question-image pairs with images in  the Karpathy test split~\cite{karpathy2017deep}
and average the scores over the questions.
The upper part of the table compares the ``training targets" for \NAME and the generic captioning model OFA-Cap.
The lower part compares the captions inferred by each captioning model with these ``gold captions".
We make several observations from the table:

\vspace{1mm}
\noindent\textbf{GPT-3 synthesized question-aware captions synthesized by GPT-3 are highly similar to the MSCOCO ground truth generic captions.} As seen in the upper part of Table~\ref{tab:captioning}, the question-aware captions are really similar to the MSCOCO ground-truth captions.

\vspace{1mm}
\noindent\textbf{\NAME achieves high CIDEr and SPICE scores using GPT-3 synthesized captions as reference.} 
The second row in the lower part compares the prompt-guided captions generated by \NAME with the GPT-3 synthesized question-aware captions. We can see that the CIDEr and SPICE scores are really high. One possible reason for the high scores is that synthesized question-aware captions are typically less diverse, shorter, and cover fewer visual entities compared with human-written general captions.
Moreover, the image captioning task becomes less ambiguous via the prompt's control, making it easier for \NAME to learn.

\vspace{1mm}
\noindent\textbf{\NAME can also generate high-quality generic captions.}
The last row shows the quality of the generic captions generated by \NAME. Users can get generic captions by prompting \NAME with the question ``what does the image describe?". All the automatic metrics show that \NAME achieves SOTA performance on COCO validation set, with even higher scores than the original OFA-Cap model.

%% file: tables/okvqa.tex
\begin{table*}[h]
\small
\centering
\caption{
\label{tab:okvqa}
Results comparison with existing systems on OK-VQA, with the image representation and the knowledge source each method uses.
GPT-3 is frozen for all methods.  The methods on top require end-to-end finetuning on OK-VQA. The methods below are fully based on in-context learning or zero-shot learning and do not require task-specific finetuning.
}
\begin{tabular}{l|c|l|c}
\toprule[1.2pt]
Method & Image Representation & Knowledge Source & Accuracy (\%)\\
\midrule
\textbf{End-to-End Finetuning}\\
Question only \cite{marino2019ok} & - & - & 14.9 \\
MUTAN \cite{marino2019okvqa} & Feature & - & 26.4 \\
BAN + KG + AUG \cite{Li2020BoostingVQ} & Feature & Wikipedia + ConceptNet & 26.7 \\
ConceptBERT \cite{garderes2020conceptbert} &  Feature & ConceptNet & 33.7\\
KRISP \cite{marino2021krisp} &  Feature & Wikipedia + ConceptNet & 38.4 \\
Vis-DPR \cite{luo2021weakly}  &  Feature  & Google Search  & 39.2 \\
MAVEx \cite{wu2022multi} & Feature & Wikipedia + ConceptNet + Google Images & 39.4  \\
TRiG \cite{Gao_2022_CVPR} & Caption + Tags + OCR & Wikipedia & 50.5 \\
KAT (Single) \cite{gui2022kat} & Caption + Tags + Feature & GPT-3 (175B) + Wikidata & 54.4   \\
KAT (Ensemble) \cite{gui2022kat} & Caption + Tags + Feature & GPT-3 (175B) + Wikidata & 54.4   \\
REVIVE (Single) \cite{Lin2022REVIVERV} & Caption + Feature & GPT-3 (175B) + Wikidata & 56.6 \\
REVIVE (Ensemble) \cite{Lin2022REVIVERV} & Caption + Feature & GPT-3 (175B) + Wikidata & 58.0 \\
\midrule
\textbf{In-Context Learning \& Zero-Shot} \\
BLIP-2 VIT-G FlanT5$_\text{XXL}$ \cite{Li2023BLIP2BL} (zero-shot) & Feature & FlanT5-XXL (11B) & 45.9 \\
PICa-Base \cite{yang2022empirical} & Caption + Tags &  GPT-3 (175B) & 43.3   \\
PICa-Full \cite{yang2022empirical} & Caption + Tags &  GPT-3 (175B) & 48.0   \\
Flamingo (80B) \cite{alayrac2022flamingo} (zero-shot) & Feature & Chinchilla (70B) & 50.6 \\
Flamingo (80B) \cite{alayrac2022flamingo} (32-shot) & Feature &  Chinchilla (70B) & 57.8 \\ 
\textbf{PromptCap + GPT-3} & Caption  & GPT-3 (175B) & \textbf{60.4} \\
\bottomrule[1.2pt]
\end{tabular}

\end{table*}

%% file: tables/aokvqa.tex
\begin{table}[h]
\small
\centering
\caption{\label{tab:aokvqa}
Results comparison with existing systems on A-OKVQA. There are two evaluations, namely multiple-choice and direct-answer. Both are measured by accuracy(\%).
}
\begin{tabular}{lcc|cc}
\toprule[1.2pt]
& \multicolumn{2}{c}{Multiple Choice} & \multicolumn{2}{c}{Direct Answer} \\
Method & val & test & val & test \\
\midrule
ClipCap \cite{schwenk2022aokvqa}  & 44.0 & 43.8  & 18.1 & 15.8 \\
Pythia \cite{jiang2018pythia} & 49.0 & 40.1  & 25.2  & 21.9 \\
ViLBERT \cite{lu2019vilbert} & 49.1 & 41.5  &  30.6  &  25.9 \\
LXMERT \cite{Tan2019LXMERTLC}  &  51.4  & 41.6 & 30.7  & 25.9 \\
KRISP \cite{marino2021krisp}  &  51.9  & 42.2  & 33.7  &  27.1 \\
GPV-2 \cite{kamath2022webly}  &  60.3  &  53.7  & 48.6  & 40.7 \\
\midrule
\textbf{PromptCap + GPT-3} & \textbf{73.2} & \textbf{73.1} & \textbf{56.3}  & \textbf{59.6}\\
\bottomrule[1.2pt]
\end{tabular}

 \vspace{-3mm}
\end{table}

%% file: tables/ablation.tex
\begin{table}[th]
\small
\centering
\caption{
Ablation on the contribution of \NAME, compared with generic captioning model OFA-Cap. The LM we use is GPT-3.
}
\begin{tabular}{l|ccc}
\toprule[1.2pt]
Captioning Model & OK-VQA & A-OKVQA & VQAv2 \\
\midrule
OFA-Cap  & 56.6 & 51.0 & 64.9 \\
\textbf{\NAME} & \textbf{60.4} & \textbf{56.3} & \textbf{74.1} \\

\bottomrule[1.2pt]
\end{tabular}
\vspace{-0.1in}
\label{tab:ablation}
\end{table}

%% file: tables/ablation_GPT3.tex
\begin{table}[th]
\small
\centering
\caption{
Ablation on the contribution of GPT-3. We measure the performance gain of using GPT-3 as the language model, compared with Flan-T5-XXL (11B).
The captioning model we use is \NAME.
}
\begin{tabular}{l|ccc}
\toprule[1.2pt]
Language Model & OK-VQA & A-OKVQA & VQAv2 \\
\midrule
Flan-T5-XXL (11B) & 42.0 & 41.5 & 70.9\\
GPT-3 (175B) & \textbf{60.4} & \textbf{56.3} & \textbf{74.1} \\
\bottomrule[1.2pt]
\end{tabular}
\label{tab:ablation_gpt3}
\end{table}

%% file: tables/ablation_prompting.tex
\begin{table}[h]
\small
\centering
\caption{
Ablation of GPT-3 prompting on OK-VQA. We experiment with different numbers of in-context examples in the input and measure the performance gain from retrieving similar in-context examples compared with random examples.
}
\begin{tabular}{l|l|cccc}
\toprule[1.2pt]
Examples & Caption & n=1 & n=4 & n=16 & n=32\\
\midrule
Random & OFA-Cap & 42.8 & 46.6 & 49.7 & 50.8\\
&\NAME & 46.5 & 50.0 & 53.1 & 55.2\\
\midrule
CLIP & OFA-Cap & 44.5 & 50.0 & 55.3 & 56.6\\
&\NAME & 48.7 & 53.3 & 58.4 & 60.4\\
\bottomrule[1.2pt]
\end{tabular}
\label{tab:prompt_ablation}
\end{table}

%% file: tables/webqa.tex
\begin{table}[h]
\small
\centering
\caption{
Results on the WebQA validation set with oracle sources on image queries.
The baselines in the upper part are fully supervised, while our methods only use 8-shot in-context learning.
}
\begin{tabular}{l|ccc}
\toprule[1.2pt]
Method & FL & Acc & FL*Acc \\
\midrule
\textbf{Fully supervised} \\
VLP + VinVL~\cite{chang2022webqa} & 47.6 & 49.6 & 27.5 \\
VLP + x101fpn~\cite{chang2022webqa} & 46.9 & 44.3 & 23.8 \\
\midrule
\textbf{8-shot in-context learning}\\
OFA-Cap + GPT-3 & 52.8 & 55.4 & 33.5\\
\textbf{\NAME + GPT-3} & \textbf{53.0} & \textbf{57.2} & \textbf{34.5}\\ 
\bottomrule[1.2pt]
\end{tabular}
\label{tab:webqa}
 \vspace{-5mm}
\end{table}

%% file: tables/captioning.tex
\begin{table}[h]
\small
\centering
\caption{Comparison of captions.
``GPT-3-Syn" are the question-aware captions synthesized by GPT-3.
``COCO-GT" are the MSCOCO ground-truth captions.
Higher scores imply higher similarities between the captions.
}
\begin{tabular}{l|l|cccc}
\toprule[1.2pt]
\multicolumn{2}{c}{Captions} & B & M & C & S\\
\midrule
\multicolumn{6}{l}{\textbf{Comparison between ``gold captions"}}\\
GPT-3-Syn & COCO-GT & 67.1 & 44.3  & 182.9 & 32.1\\
\midrule
\multicolumn{6}{l}{\textbf{Inferenced captions vs. ``gold captions"}}\\
OFA-Cap &  GPT-3-Syn & 26.2 &  25.3  & 231.0  & 40.2\\
\textbf{\NAME} & GPT-3-Syn & \textbf{33.0}  & \textbf{29.7}  & \textbf{307.1} & \textbf{47.3}\\
OFA-Cap & COCO-GT & 44.5 & 30.9 & 147.9 & 24.6\\
\textbf{\NAME} & COCO-GT & \textbf{45.4}  & \textbf{31.6}  & \textbf{150.1} & \textbf{25.2}\\
\bottomrule[1.2pt]
\end{tabular}
\label{tab:captioning}

 \vspace{-3mm}
\end{table}

%% file: text/discussion.tex
\section{Limitations and Broader Impact}

\begin{figure}[ht]
\centering
  \includegraphics[width=0.45\textwidth]{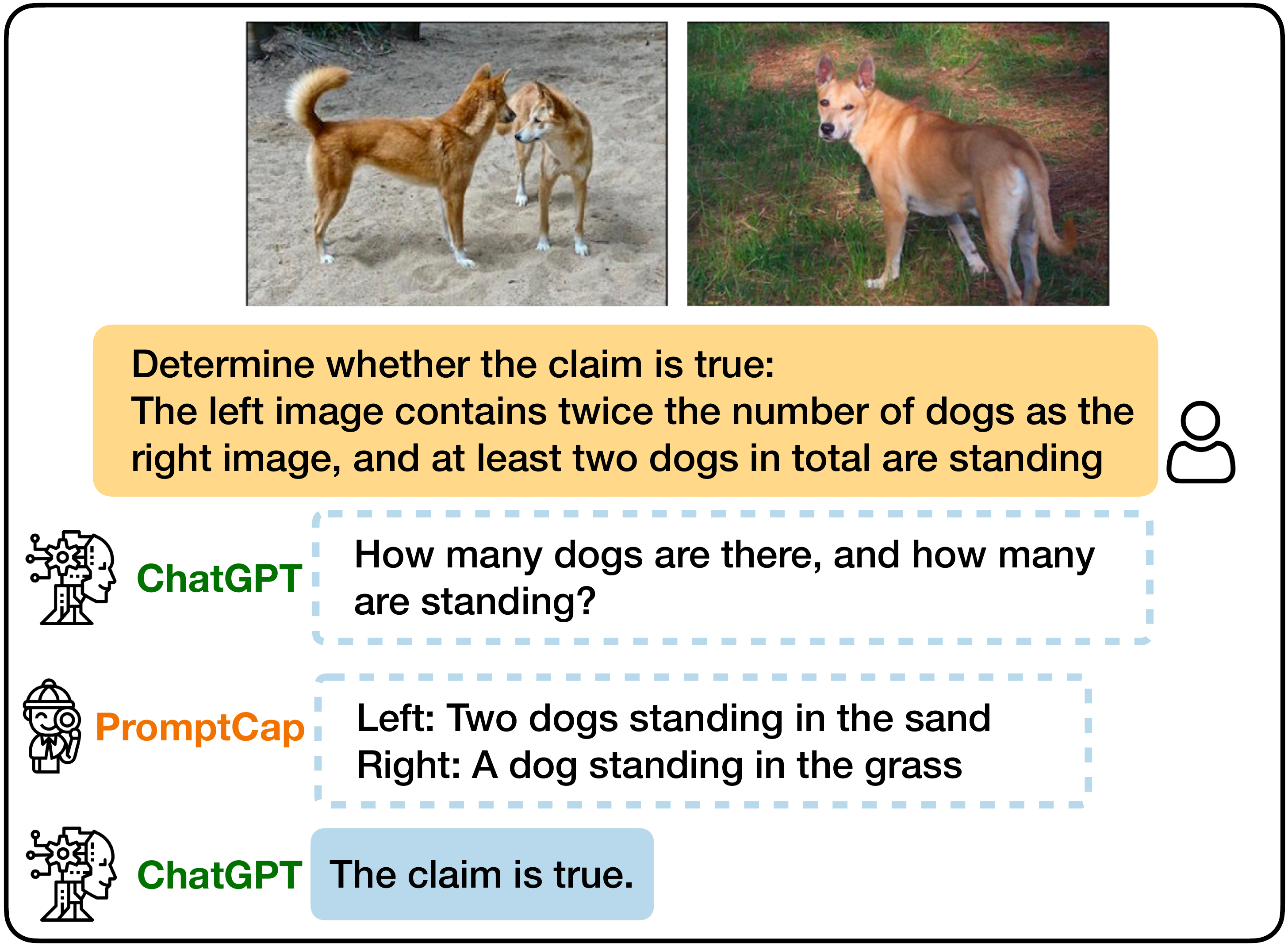}
  \caption{
Demo of solving the NLVR2 task with off-the-shelf \NAME and ChatGPT via an interpretable reasoning process.
}
  \vspace{-0.1in}
  \label{fig:extension}
\end{figure}

One limitation is that the current \NAME only focuses on knowledge-based VQA tasks. 
\NAME can be extended to other vision-language tasks beyond VQA. Figure~\ref{fig:extension} shows an example of solving NLVR2~\cite{suhr2019orpus} via a series of vision and reasoning steps between \NAME and ChatGPT.
Future work may scale up \NAME training with more diverse tasks and instructions, and explore broader applications of \NAME beyond VQA.

Another limitation is that images contain information that cannot be abstracted as text. While \NAME has demonstrated promising results in bridging the gap between LMs and images, it is important to recognize its limitations and use it in conjunction with other methods to ensure a comprehensive understanding of visual data.

%% file: text/conclusion.tex
\section{Conclusion}
We present \NAME, a novel question-aware captioning model that can be controlled via a natural language prompt.
To train this captioning model with no extra annotation, we devise an efficient pipeline for synthesizing and filtering training examples via GPT-3.
We demonstrate the effectiveness of \NAME on knowledge-based VQA tasks.
Our system achieves state-of-the-art performance on OK-VQA and A-OKVQA.
Ablations show that \NAME is giving consistent gains over generic captions.
Furthermore, we investigate \NAME's generalization ability on WebQA.
\NAME works as a simple
and general module for converting question-related visual information into text.


%% file: text/appendix.tex
\section{More Implementation Details}

The pre-trained OFA~\cite{wang2022ofa} is trained on about 20M image-text pairs. We further fine-tune OFA using examples synthesized from VQAv2~\cite{goyal2017making} on COCO 2014 training set, which contains 443757 examples. Notice that we haven't used COCO 2014 validation set to avoid data leakage.

For model training, the most important details are given in Section~\ref{sec:promptcap}. 
We finetune \NAME on the synthesized dataset for 2 epochs, which takes about 10 hours on 4 NVIDIA A40 GPUs. 
For all the experiments that use random in-context examples for GPT-3, we report the average result of 3 runs.

\section{Common Q \& A}
\label{sec:appendix:qa}

\paragraph{Compared with OFA, does the performance gain of \NAME come from more training data?}
No. As discussed in \S\ref{sec:exp:ablation}, OFA is a large-scale vision-language model pre-trained on 20M image-text pairs and 20 vision-language tasks, including many VQA tasks. \NAME is fine-tuned on VQAv2~\cite{goyal2017making}, which is included in OFA training data. The performance gain comes from the idea of \textbf{controlling image captions with natural language instructions.} Compared with OFA, we synthesize VQAv2 into prompt-guided captioning tasks via GPT-3 and use it to further fine-tune OFA. This way, OFA gains the ability of prompt-guided captioning with \textbf{the same amount of annotated data}, leading to significant performance gains.

\paragraph{Can \NAME do generic captioning?}
Yes. We find that prompting \NAME with the question ``what does the image describe?" leads to high-quality generic captions. The CIDEr is 150.1 on the COCO validation set in the setting in \S\ref{sec:appendix:captioning}.

\section{Prompt Examples}
\label{appendix:prompt}

In this section, we show examples of the prompts we used for prompting GPT-3.

\subsection{Prompt for example synthesis}
\label{appendix:few shot}
Below is the full version of the prompt for training example synthesis with GPT-3.
We formulate the task into a summarization tasks, and include the human-written examples in the prompt. The question-answer pair synthesized is at the end of the prompt, and GPT-3 generates the prompt-guided caption example.

\begin{tiny}
\begin{lstlisting}[breaklines]
Summarize the context to help answer the question

Original contexts: A very clean and well decorated empty bathroom. A blue and white bathroom with butterfly themed wall tiles. A bathroom with a border of butterflies and blue paint on the walls above it.
Question: Is the sink full of water?
Answer: no
Summary: A bathroom with an empty sink.

Original contexts: Several metal balls sit in the sand near a group of people.. People standing around many silver round balls on the ground.. Silver balls are lined up in the sand as people mill about in the background.. Silver balls on sand with people walking around. . silver balls laying on the ground around a smaller red ball.
Question: What color are the round objects?
Answer: silver
Summary: People standing around many silver round balls on the ground.

Original contexts: A kitchen cart stands in the middle of a kitchen with wooden cupboards.. A bright kitchen with hardwood floors and wooden cupboards.. a kitchen that is fully furnished with a cart in the center of it.. The kitchen is brightly lit from the window.. Brightly colored kitchen with wooden floors, large images on wall, and small island.
Question: What is the source of light in this picture?
Answer: sun
Summary: A bright kitchen lit by sunlight.

Original contexts: An empty kitchen with white and black appliances.. A refrigerator and stove are in a small kitchen area. . Small kitchen in a personal home with dual sinks.. A small kitchen with sink, stove and refrigerator.. A small kitchen with several appliances and cookware.
Question: How many cabinets in this room?
Answer: 4
Summary: A small kitchen with 4 cabinets.

Original contexts: Green tiled backsplash highlighted by low overhead lighting.. A kitchen counter is illuminated by a hood light. A kitchen sink next to an empty counter with a tiled wall.. A back splash is added to the wall in the kitchen.. A picture of a sink top with dim lighting.
Question: What material is the backsplash made of?
Answer: tile
Summary: Green tiled backsplash highlighted by low overhead lighting.

Original contexts: A graffiti-ed stop sign across the street from a red car . A vandalized stop sign and a red beetle on the road. A red stop sign with a Bush bumper sticker under the word stop.. A stop sign that has been vandalized is pictured in front of a parked car.. A street sign modified to read stop bush.
Question: What season is it in this photo?
Answer: summer
Summary: A stop sign and a car on a street in summer.

Original contexts: Lady carrying a purse walking along side a man.. A city sidewalk is lined with lamp posts. A man and a woman stand on the sidewalk lined with street lights.. A city sidewalk with storefronts on the right.. Two people leaving a building to walk down the street.
Question: Which item in this picture helps people see after dark?
Answer: streetlight
Summary: A city sidewalk lit by streetlight.

Original contexts: A sink and a toilet inside a small bathroom.. White pedestal sink and toilet located in a poorly lit bathroom.. Clean indoor bathroom with tiled floor and good lighting.. a bathroom with toilet and sink and blue wall. a blue bathroom with a sink and toilet.
Question: How many rolls of toilet paper are on the shelves above the toilet?
Answer: 0
Summary: A bathroom with a toilet and a sink. There is no toile paper on the shelves.

Original contexts: A couple enjoying beverages and a snack on a sunny day. Showing a  doughnut while holding drinks near a car.. A man and woman sharing apple cider and a doughnut. Two people are standing in front of an open car trunk holding drinks and a doughnut. . A man and a woman eating donuts and having drinks.. A man holding beer and a woman holding a pastry and beer.
Question: How do we know this guy is not likely to have packed a razor?
Answer: has beard
Summary: A man with beard and a woman are eating donuts and having drinks.

Original contexts: Woman riding a bicycle down an empty street.. A woman in green is riding a bike.. a woman wearing a bright green sweater riding a bicycle. A woman on a bicycle is going down the small town street.. A woman bikes down a one way street.
Question: What kind of fruit is the helmet supposed to be?
Answer: watermelon
Summary: A woman with a watermelon style helmet riding a bicycle.

Original contexts: A panoramic view of a kitchen and all of its appliances. A panoramic photo of a kitchen and dining room A wide angle view of the kitchen work area multiple photos of a brown and white kitchen.  A kitchen that has a checkered patterned floor and white cabinets.
Question: Is the counter curved?
Answer: no
Summary: A photo of a kitchen with a counter that is not curved.

Original contexts: A woman is walking a dog in the city.. A woman and her dog walking down a sidewalk next to a fence with some flowers. . A woman walking her dog on the sidewalk.. A woman walks her dog along a city street.. A woman walks her dog on a city sidewalk.
Question: What color vehicle is closest to the mailbox?
Answer: silver
Summary: A silver vehicle next to a mailbox on the sidewalk.

Original contexts: some pancakes cover with bananas, nuts, and some whipped cream . Two pancakes on top of a white plate covered in whipped cream, nuts and a banana.. Pancakes with bananas, nuts and cream, covered in syrup. . Pancakes topped with bananas, whipped cream and walnuts.. Pancakes topped with bananas, nuts, and ice cream.
Question: What restaurant was this dish cooked at?
Answer: ihop
Summary: Pancakes with banans, nuts, and cream, cooked at ihop.

Original contexts: The two people are walking down the beach.. Two people carrying surf boards on a beach.. Two teenagers at a white sanded beach with surfboards.. A couple at the beach walking with their surf boards.. A guy and a girl are walking on the beach holding surfboards.
Question: What is on the man's head?
Answer: hat
Summary: A man and a woman walking on the beach with surfboards. The man is wearing a hat.

Original contexts: A sink and a toilet inside a small bathroom.. White pedestal sink and toilet located in a poorly lit bathroom.. Clean indoor bathroom with tiled floor and good lighting.. a bathroom with toilet and sink and blue wall. a blue bathroom with a sink and toilet.
Question: Is there natural light in this photo?
Answer: no
Summary: A photo of a small bathroom in artificial light.

Original contexts: Fog is in the air at an intersection with several traffic lights.. An intersection during a cold and foggy night.. Empty fog covered streets in the night amongst traffic lights.. City street at night with several stop lights.. It is a foggy night by a traffic light.
Question: Which direction is okay to go?
Answer: straight
Summary: A traffic light in a foggy night, showing it is okay to go straight.

Original contexts: A graffiti-ed stop sign across the street from a red car . A vandalized stop sign and a red beetle on the road. A red stop sign with a Bush bumper sticker under the word stop.. A stop sign that has been vandalized is pictured in front of a parked car.. A street sign modified to read stop bush.
Question: What color is the car driving north?
Answer: red
Summary: A stop sign and a red car driving north.

Original contexts: A man in a wheelchair and another sitting on a bench that is overlooking the water.. Two people sitting on dock looking at the ocean.. Two older people sitting down in front of a beach.. An old couple at the beach during the day.. A person on a bench, and one on a wheelchair sitting by a seawall looking out toward the ocean.
Question: What is the person on the left sitting on?
Answer: bench
Summary: A person sit on a bench on the left, and another sitting in a wheelchair on the right, all looking at the ocean.

Original contexts: A parked motor scooter sitting next to a bicycle.. A picture of a motorbike and two pedal bicycles.. A motor scooter that has an advertisment on the back next to a bicycle.. A grey moped parked by building next to a bicycle.. a motor bike parked next to a bike by a building.
Question: Which model of bike is shown in this picture?
Answer: vespa
Summary: A vespa bike parking next to a bicycle.

Original contexts: People standing around a park bench next to a bicycle.. A group of women are describing a new setup for a building plan. a group of people in a field of grass near a building. Several people standing in an area with picnic tables looking at a board.. A woman giving a presentation in a park.
Question: What is the woman in the blue jacket standing on?
Answer: bench
Summary: A woman in blue jacket standing on a bench, with a group of people around her.

Original contexts: A orange tabby cat laying down on a black car. An orange cat laying on the hood on a car.. A cat sits on top of a black car.. A cat that is sitting on top of a black car.. A yellow cat sleeping on the hood of a black car parked in the garage.
Question: What brand of car is this?
Answer: subaru
Summary: An orange cat laying on top of a black subaru.

Original contexts: A bicycle parked in front of a building next to a pile of garbage.. Black and white photograph of a homeless person under their many belongings. Two people huddle on a bench under their belongings.. A homeless person is bundled within a pile of belongings..  an image of two homeless people laying under debris on a bench
Question: How is the bike affixed to the pole?
Answer: chain
Summary:
-----Prompt Ends Here-----
LM completion: A bicycle chained to a pole, with a pile of garbage next to it.
\end{lstlisting}
\end{tiny}

\subsection{GPT-3 In-Context Learning for VQA}

Here we show an example of solving VQA with GPT-3 in-context learning on OK-VQA~\cite{marino2019okvqa}.
We use the same prompt template as PICa~\cite{yang2022empirical}. This example contains 8 closest examples retrieved from the training set.

\begin{tiny}
\begin{lstlisting}
Please answer the question according to the above context.

===
Context: a bowl of broccoli and lemon slices on a table 
===
Q: How do you make that?
A: steam

===
Context: an orange tree with oranges behind a fence 
===
Q: What fruit is that?
A: orange

===
Context: a bowl of oranges and limes on a table 
===
Q: What types of fruit are these?
A: orange and lime

===
Context: two oranges and a green leaf on a white table 
===
Q: What fruits are those?
A: orange

===
Context: a basket of oranges sitting on a wooden table 
===
Q: What family of fruits is shown?
A: citrus

===
Context: a green apple sitting on top of a bunch of bananas 
===
Q: What type of fruit is this?
A: apple

===
Context: a bowl filled with oranges sitting on top of a table 
===
Q: Where can this fruit be found?
A: tree

===
Context: an orange cut in half on a white plate 
===
Q: What fruit is this?
A: orange

===
Context: a glass bowl filled with fruit on top of a table 
===
Q: What is the fruit bowl made of?
A:
-----Prompt Ends Here-----
LM completion: glass
\end{lstlisting}
\end{tiny}